\documentclass[letterpaper, 10 pt, conference]{ieeeconf}  

\usepackage{cite}
\usepackage{amsmath,amsfonts}
\usepackage{graphicx}
\usepackage{textcomp}
\usepackage{xcolor}

\usepackage{booktabs} 
\usepackage{verbatim}
\usepackage{graphics} 
\usepackage{epsfig} 
\usepackage{soul}
\usepackage{epstopdf}
\usepackage{subfigure}
\usepackage{balance}
\usepackage{comment}
\usepackage{booktabs}
\usepackage{amsmath}
\usepackage{amsfonts}
\usepackage{bm}
\usepackage{colortbl}
\usepackage{tabularx}
\usepackage{color}
\usepackage{xspace}
\usepackage{graphicx}    
\usepackage{url}         
\usepackage{algorithm}
\usepackage{algpseudocode}
\usepackage{placeins}
\usepackage{multirow}
\let\labelindent\relax
\usepackage{enumitem}
\usepackage{leading}
\usepackage{pgfplots}
\usepackage{tikz}
\usepackage{xcolor}

\usepackage{amsmath}

\IEEEoverridecommandlockouts                              

\title{\LARGE \bf
Do We Need iPhone Moment or Xiaomi Moment for Robots? \\
Design of Affordable Home Robots for Health Monitoring 
}

\author{Bo Wei$^{1*}$, Yaya Bian$^{2}$, Mingcen Gao$^{3}$
\thanks{$^{1}$Bo Wei is the corresponding author and with School of Computing,
        Newcastle University, Newcastle, UK.
        {\tt\small bo.wei@newcastle.ac.uk}; %
 $^{2}$Yaya Bian is with Newcastle Business School,
        Northumbria University, Newcastle, UK.
        {\tt\small yaya2.bian@northumbria.ac.uk}; %
$^{3}$Mingcen Gao is working in Google, San Francisco, USA
        {\tt\small gaomingcen@gmail.com}}%
}

\begin{document}

\maketitle
\thispagestyle{empty}
\pagestyle{empty}



\begin{abstract}
In this paper, we study cost-effective home robot solutions which are designed for home health monitoring. The recent advancements in Artificial Intelligence (AI) have significantly advanced the capabilities of the robots, enabling them to better and efficiently understand and interact with their surroundings. The most common robots currently used in homes are toy robots and cleaning robots. While these are relatively affordable, their functionalities are very limited. On the other hand, humanoid and quadruped robots offer more sophisticated features and capabilities, albeit at a much higher cost. Another category is educational robots, which provide educators with the flexibility to attach various sensors and integrate different design methods with the integrated operating systems. However, the challenge still exists in bridging the gap between affordability and functionality. Our research aims to address this by exploring the potential of developing advanced yet affordable and accessible robots for home robots, aiming for health monitoring, by using edge computing techniques and taking advantage of existing computing resources for home robots, such as mobile phones. 

\end{abstract}

\section{INTRODUCTION}\label{sec:intro}
With over 1 billion people aged 60 years and older in 2019, this number is projected to reach 2.1 billion by 2050 \cite{improving2024}. This demographic shift necessitates of health monitoring that play an important role in ensuring the well-being of older individuals. Health monitoring helps timely detection of health issues, promotes preventive measures, and thus enhances overall quality of life.


While care workers at care homes and elderly caregivers at home are common solutions for senior care, a significant barrier exists, i.e., cost. Accessing professional care services is not affordable for many older adults from non-wealthy families. Therefore, it becomes important to explore alternative solutions that provide affordable and accessible care for the ageing population. 
%
In the home health monitoring area, robots emerge as a promising option.
There are several kinds of common robots. 
Toy robots and cleaning bots, while relatively affordable, remain limited in their functionalities \cite{irobot}. On the other hand, humanoid and quadruped robots, developed or being developed by industry giants like Boston Dynamics \cite{dynamics2024}, Tesla \cite{tesla2024}, and Nvidia \cite{nvidia2024} 
offer advanced features and capabilities. However, their sophistication comes at a higher cost. Another emerging category for daily use is educational robots \cite{ros}.
These educational robots serve as effective tools for teaching robotics and AI concepts. Despite these advancements in these popular robots, the challenge persists in finding a balance between affordability and functionality, which is the key to ensuring that cutting-edge technology remains accessible to all. We envision a future where robots seamlessly collaborate with humans in home health monitoring, so bridging this affordability-functionality gap becomes essential. 



Our research aims to address this by exploring the potential of developing advanced yet affordable robots for home health monitoring by using edge computing techniques.
The integration of AI with robotics can innovate home healthcare, but it requires extensive computational units, which increases robot costs. The use of edge computing techniques will let robots use the local existing computational resources, such as mobile phones, which reduces the cost of home robots. We will also consider a cost-effective supply chain to further reduce the cost.

The rest part of this paper is organised as follows. We show the related works in Section \ref{sec:related} and the details of our design in Section \ref{sec:system}. The paper is finally concluded in Section \ref{sec:conclusion}.
\vspace{-5mm}
\section{RELATED WORKS}\label{sec:related}
\vspace{-2mm}
Toy robots and cleaning robots are very common and affordable \cite{irobot}. These robots are equipped with cost-effective sensors with low specification, such as inertial measurement unit (IMU), ultrasound sensors, cameras, etc. 
Their simple functionalities, such as responding to voice commands or performing basic movements, contribute to mental stimulation and emotional well-being. Meanwhile, cleaning robots can sweep floors and vacuum carpets. However, their limitation is the lack of sophisticated humanoid counterparts.
The development of humanoid and quadruped robots represents a significant leap forward. These robots, developed by industry leaders such as Boston Dynamics \cite{dynamics2024}, Tesla \cite{tesla2024}, and Nvidia \cite{nvidia2024}, offer a blend of sophistication and versatility with autonomous navigation, sensor fusion, real-time data collection, and manipulation. With high-specification sensors, such as Lidar, laser, etc,  providing rich information on the surrounding environment, these robots can traverse and perform complicated tasks.
The limitation for these robots is the cost, where such one robot costs at least more than \$10,000. 
Another common type is educational robots \cite{ros}. Educators and students are able to attach sensors and implement different algorithms by using them.
They serve as effective tools for teaching and learning robotics concepts as well as AI concepts, but they are not as capable as humanoid and quadruped robots and are not affordable. 



%

\begin{figure}[!t]
\vspace{3mm}	
   \centering
    \includegraphics[width=.35\textwidth]{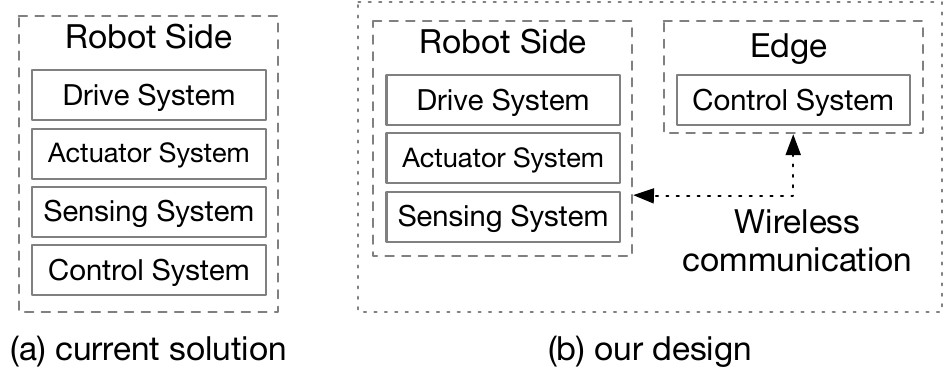}
    \vspace{-3mm}
    \caption{Current solution and our design}\label{fig:overview}
    \vspace{-8mm}
\end{figure}
\section{SYSTEM OVERVIEW}\label{sec:system}
As shown in Figure \ref{fig:overview}(a), a robot is composed of four parts: (1) Drive system: the drive system provides the necessary motion for the robot, including components like motors, wheels, tracks, or legs that allow the robot to move through its environment. 
(2) Actuator system: the actuator system (e.g. servos, hydraulic cylinders, or pneumatic pistons) translates electrical signals into physical movement and executes specific tasks or actions which enable the robot to manipulate objects, lift weights, or perform precise movements.
(3) Sensing system: the sensing system equips the robot with perception capabilities. Sensors gather information from and understand the surroundings and environment. Common sensors include cameras, lidar, ultrasonic sensors, and infrared detectors. By processing sensory data, the robot can detect obstacles, measure distances, and recognise objects. (4) Control system: the control system processes input information from sensors, makes decisions, and generates output commands for the actuator and drive systems. 
 \vspace{-1mm}
\section{Edge Enabled Control System for Affordable Robots Design }\label{sec:method}
 \vspace{-1mm}
In this section, we will give a detailed design to reduce the cost of home robots.
Traditionally, robots carry their control systems onboard, often comprising powerful computing units to enable complex algorithms. However, the control system contributes the majority of the overall cost, limiting the affordability of robotic solutions. To address this challenge, we move the control system from the robot side to the edge side and use existing computing units, such as our widely used mobile phone, as the control system (shown in Figure \ref{fig:overview}(b)). 
Only cheap microcontrollers, along with communication modules, are kept and integrated into the robot to transfer data collected from sensors to the edge side. Although this paper focuses on the discussion of using mobile phones, local computers or other existing edge computing units can also be used. Carried mobile phones were used as robot computing units in the solution \cite{oros2013smartphone}, but communication, task scheduling, and complicated computation tasks have not been fully considered when an edge device is used.
Specifically, we will make the following considerations for our design:

\noindent \textbf{Mobile Phones as Control System:} These common edge computing units are equipped with powerful processors, sufficient memory, and communication capabilities. By re-purposing these widely used edge computing units as control systems of robots, we use their computational power without incurring additional costs. 
The sensors in a robot still collect information about its surrounding environment. Instead of processing the data onboard, the robot transfers them to the edge side, i.e., the mobile phone in our context. Communication protocols, such as Wi-Fi, and efficient task scheduling can facilitate efficient data exchange for real-time decision-making. 

\noindent \textbf{Edge Intelligence and Decision-Making:} The mobile phone's processing capabilities enable edge intelligence. Algorithms running on the edge analyse sensor data, make decisions, and generate control signals for robots, including tasks like obstacle avoidance, path planning, and manipulation. Many mobile phones have dedicated neural processing units to realise on-device inference for complicated machine learning models, such as Neural Engines in iPhone \cite{banerjee2018microarchitectural}, Tensor processor in Google Pixel \cite{ignatov2019ai}, etc. 

\noindent \textbf{Minimalistic Robot Hardware:} To make the robot more affordable, we reduce its onboard components. Only microcontrollers and essential communication modules are integrated directly into the robot to communicate with the edge. The drive system, the actuator system and the sensing system are explicitly designed without adding extra unnecessary functions. 

\noindent \textbf{Leveraging the Supply Chain:} 
By optimising supply chain processes, the affordability of robotic solutions can be significantly enhanced, which includes effective sourcing and procurement practices. Negotiating favourable terms with suppliers and manufacturers helps secure designed robots at competitive prices. Manufacturers can also buy components in bulk. This reduces per-unit costs and ensures a steady supply of affordable materials. Diversifying sources can also ensure continuity even if one supplier faces disruptions, which can also prevent production delays.



\vspace{-1mm}
\section{CONCLUSIONS}\label{sec:conclusion}
\vspace{-1.5mm}
In conclusion, 
we aim to achieve the design of affordable home robots by moving the control system to the edge side, i.e., a mobile phone and realise edge-enabled decision-making, as well as limiting the components to the minimal requirement and leveraging effective supply chain. Affordable robot design represents a forward-thinking approach to adapting our societies to meet the challenges of an ageing demographic, ensuring that everyone can enjoy a long, healthy life in an age-friendly environment.
\vspace{-2mm}











\bibliographystyle{IEEEtran}
\bibliography{sigproc}

\end{document}